\documentclass{article}

\PassOptionsToPackage{numbers,compress}{natbib}
    \usepackage[final]{neurips_2024}

\usepackage[utf8]{inputenc} % allow utf-8 input
\usepackage[T1]{fontenc}    % use 8-bit T1 fonts
\usepackage[hidelinks]{hyperref}       % hyperlinks
\usepackage{url}            % simple URL typesetting
\usepackage{booktabs}       % professional-quality tables
\usepackage{amsfonts}       % blackboard math symbols
\usepackage{nicefrac}       % compact symbols for 1/2, etc.
\usepackage{microtype}      % microtypography
\usepackage{xcolor}         % colors
\usepackage{subfig}
\usepackage{graphicx}
\usepackage{amsmath}
\usepackage{gensymb}

\title{Continuous latent representations for modeling precipitation with deep learning}

\author{%
  Gokul Radhakrishnan \\
  Verisk Analytics \\
  Hyderabad, IND 500081 \\
  \texttt{gokul.r@verisk.com} \\
  \And
  Rahul Sundar \\
  Verisk Analytics \\
  Hyderabad, IND 500081 \\
  \texttt{rahulsundar@verisk.com} \\
  \And
  Nishant Parashar \\
  Verisk Analytics \\
  Hyderabad, IND 500081 \\
  \texttt{nparashar@verisk.com} \\
  \And
  Antoine Blanchard \\
  Verisk Analytics \\
  Boston, MA 02115 \\
  \texttt{ablanchard@verisk.com} \\
  \And
  Daiwei Wang \\
  Verisk Analytics \\
  Boston, MA 02115 \\
  \texttt{dwang@verisk.com} \\
  \And
  Boyko Dodov \\
  Verisk Analytics \\
  Boston, MA 02115 \\
  \texttt{bdodov@verisk.com} \\
}
\begin{document}
% \linenumbers

\maketitle

\begin{abstract}
The sparse and spatio-temporally discontinuous nature of precipitation data presents significant challenges for simulation and statistical processing for bias correction and downscaling. These include incorrect representation of intermittency and extreme values (critical for hydrology applications), Gibbs phenomenon upon regridding, and lack of fine scales details. To address these challenges, a common approach is to transform the precipitation variable nonlinearly into one that is more malleable. In this work, we explore how deep learning can be used to generate a smooth, spatio-temporally continuous variable as a proxy for simulation of precipitation data. We develop a normally distributed field called pseudo-precipitation (PP) as an alternative for simulating precipitation. The practical applicability of this variable is investigated by applying it for downscaling precipitation from \(1\degree\) (\(\sim\) 100 km) to \(0.25\degree\) (\(\sim\) 25 km).
\end{abstract}

\section{Introduction}
\label{sec:intro}
Precipitation is a key driver of the Earth's hydrological cycle, making its accurate modeling crucial for studying atmospheric processes. Accurate estimation of precipitation is vital for various human activities, such as transportation and agriculture. Unlike smoother meteorological variables such as temperature, water vapor, and wind speed, precipitation data is sparse and exhibits significant spatial variability. Despite major advancements in numerical weather prediction (NWP) and global circulation models (GCMs), these models still face challenges in accurately predicting extreme precipitation events, like heavy rainfall, due to limitations in resolution and parameterization. These models are further constrained by high computational demands of simulating global climate. 

Precipitation data presents several inherent complexities that make its post processing particularly challenging. Precipitation has high spatio-temporal variability, resulting in vast regions with zero values interspersed with sporadic positive values that can increase exponentially in magnitude \cite{tapiador2019precipitation}. The low frequency of extreme precipitation events adds to the complexity \cite{shi2017deep,scheuerer2015statistical}. Moreover, both precipitation and the various multi-scale factors contributing to its formation display non-normal and nonlinear behaviors.

These challenges are particularly evident in downstream applications such as statistical post-processing \cite{zhang2023statistical}, downscaling \cite{maraun2010precipitation, kumar2023modern, sachindra2018statistical}, nowcasting \cite{shi2017deep}, and forecasting \cite{li2022using, ridwan2021rainfall}. Various research groups have utilized statistical methods to address the complexities of precipitation data, especially in bias correction \cite{wang2023customized, le2020application, wang2022deep}. The statistical post-processing of simulated precipitation from NWP models lack proper consideration of a number of moisture-related properties of non-precipitating members of the ensemble that likely have discriminating information on the calibration forecasts. This issue is more pronounced when the ensemble forecast is dry-biased, making the statistical adjustment process more complicated. To address this issue, Yuan \textit{et al.} \cite{Yuan2019} proposed a statistically continuous variable called pseudo-precipitation obtained after blending precipitation and integrated vapour deficit (IVD) together. 

To achieve a consistent representation for precipitation while preserving its key characteristics, we propose using machine learning for generating a pseudo-precipitation field. For transforming total precipitation (TP) into a spatiotemporally continuous field, we use  vertically integrated moisture divergence (VIMD) \cite{banacos2005use}. VIMD contains relevant information pertaining to decrease (divergence) or increase (convergence) of moisture within a vertical column of air. Unlike IVD, VIMD can take both negative and postive values and its spatial correlation structure is similar to TP. This can potentially enable more effective blending, specifically at point of discontinuity through deep learning techniques, as detailed in Section \ref{sec:methods}. Further, we perform the blending of our pseudo-precipitation field targeted towards a symmetric Gaussian distribution. The smoother Gaussian blending makes precipitation data more manageable for analysis, enhancing the coherence and accuracy of post-processing models. Additionally, it offers improved physical consistency by representing the processes driving precipitation patterns and facilitate the integration of precipitation with other climate variables. To assess the practical applicability of pseudo-precipitation, we evaluate its efficacy for the downscaling task using generative deep learning \cite{rampal2024robust, karras2022elucidating}. 

\section{Methods}
\label{sec:methods}
\paragraph{Pseudo-precipitation (PP)}
We define PP as a smooth, Gaussian field generated by blending TP and VIMD, inspired by the work of Yuan \textit{et al.} In \cite{Yuan2019}, IVD was used to generate the PP field. IVD represents the difference between actual and saturation specific humidity, integrated throughout the troposphere. IVD measures dryness of the atmospheric column and hence is always negative. The authors defined PP as being equal to precipitation when precipitation is positive while the non-positive values are replaced by IVD after suitable transformation to ensure continuity of the blended field. 

In this work, we use VIMD as a replacement for IVD. VIMD is defined as the vertical integral of the moisture flux for a column of air extending from the surface of the Earth to the top of the atmosphere. Its horizontal divergence is the rate of moisture spreading outward from a point, per square metre. Positive values indicate moisture divergence (dry conditions) and negative values indicate moisture convergence (potential condensation). VIMD's spatial correlation structure closely resembles that of TP, making it a suitable candidate for blending with TP. To ensure seamless integration of VIMD and TP, we blend them into a Gaussian distribution as symmetric distributions are preferred for statistical processing.  Additionally VIMD is a native ERA5 variable along with TP facilitating ease in analysis.\footnote[1]{https://codes.ecmwf.int/grib/param-db/213}

\begin{figure}[h]
  \centering
  \includegraphics[width=1.0\textwidth]{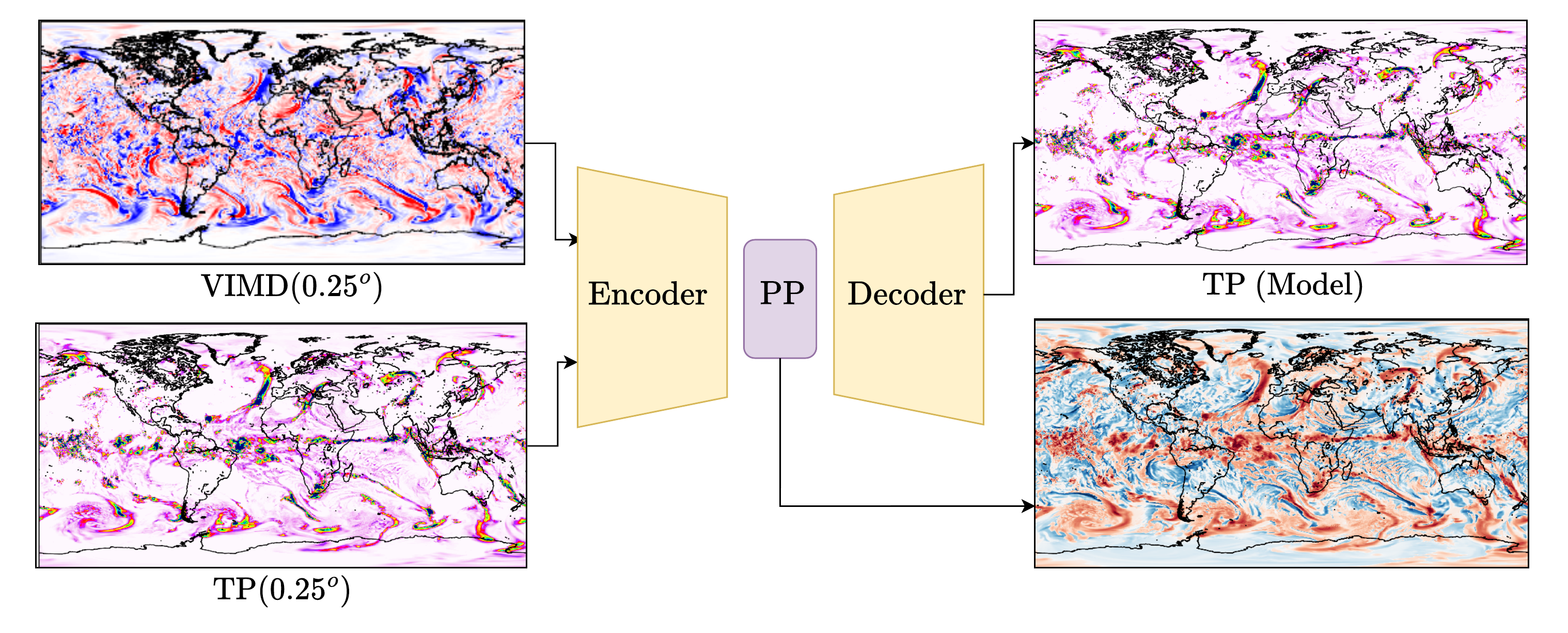}
  \caption{Schematic of the mapping model for PP}
  \label{fig:schematics}
\end{figure}

Our approach utilizes a fully connected encoder-decoder framework (figure \ref{fig:schematics}), trained on point-wise global ERA5 reanalysis data \cite{hersbach2020era5}. The encoder blends TP and VIMD into a Gaussian-distributed field, pseudo-precipitation (PP). A quantile loss is used to align the distribution of PP with that of a standard normal distribution. The decoder then reconstructs TP from PP. Compared to Yuan \textit{et al.} \cite{Yuan2019}, this neural network framework offers a more flexible and expressive way to parameterize the blended field, while also enabling the decoding of precipitation from the blended field.

\begin{figure}[t]
  \centering
  \includegraphics[width=1.02\textwidth]{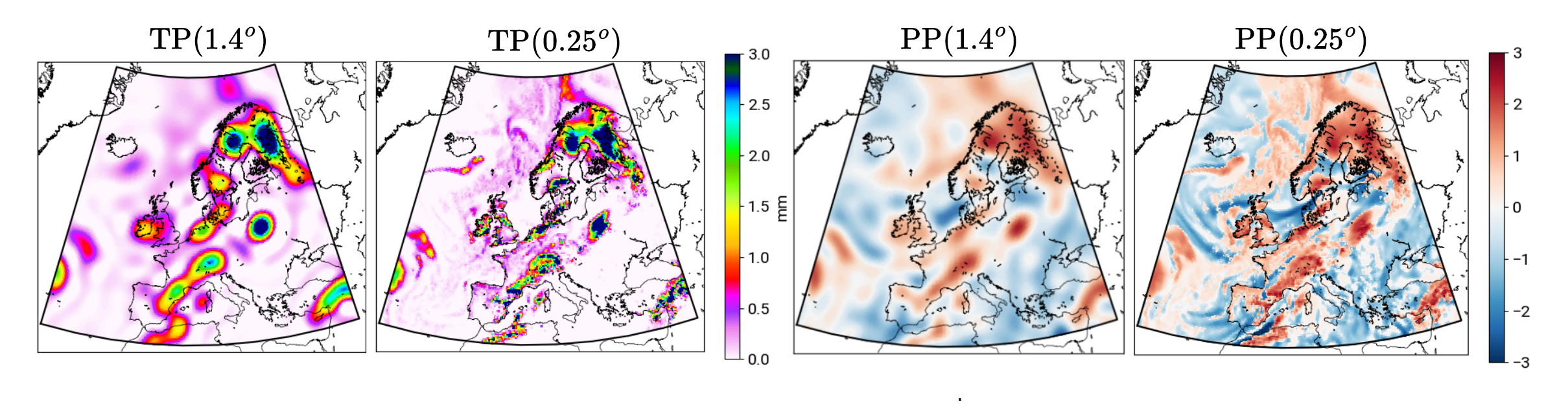}
  \caption{Low and high resolution pairs of TP (left) and PP (right).}
  \label{fig:wavelet}
\end{figure}

\paragraph{Limitations of TP for downscaling}
The overarching limitations of TP are outlined in Section \ref{sec:intro}. For training a supervised model for downscaling, we generate samples at coarse resolution. A common method for this is spectral truncation (based on spherical harmonics or spherical wavelets) \cite{antoine2002wavelets, mcewen2018localisation} as it is a robust framework for scale separation. %This allows direct applicability of the trained downscaling models on debiased GCM simulations. However,% 
Applying spectral truncation on TP leads to non-physical artifacts, particularly oscillations or "ringing" near sharp edges or discontinuities \cite{kelly1996gibbs} due to Gibbs phenomena, as illustrated in figure \ref{fig:wavelet}a for spherical wavelet transforms. In contrast, the proposed pseudo-precipitation field does not suffer from this issue (figure \ref{fig:wavelet}b).

\paragraph{Overview of the downscaling framework} 

To demonstrate the benefits of PP, we evaluate its efficacy in the context of downscaling. First, we generate paired low-resolution and high-resolution PP data from ERA5 reanalysis.  The high-resolution data is at ERA5’s native \(0.25\degree\) (\(\sim 25\) km) resolution. The low-resolution data is generated by spherical wavelet transforms of the high-resolution data, producing band-limited fields at a resolution of \(1.4\degree\)  (\(\sim  70\) km), as shown in figure \ref{fig:wavelet}. Our downscaling framework is described in Sundar \textit{et al.} \cite{taudiff2024}; which integrates a spatio-temporal model, SimVP \cite{gao2022simvp, tan2023temporal}, with a diffusion model \cite{dhariwal2021diffusion, karras2022elucidating, mardani2024residual}. Once the downscaling model is trained on PP, we decode TP at the target resolution using the decoder in figure \ref{fig:schematics}. Downscaled, decoded TP is used for investigating the overall performance of our model.

\begin{figure}[h]
  \centering
  \includegraphics[width=1.02\textwidth]{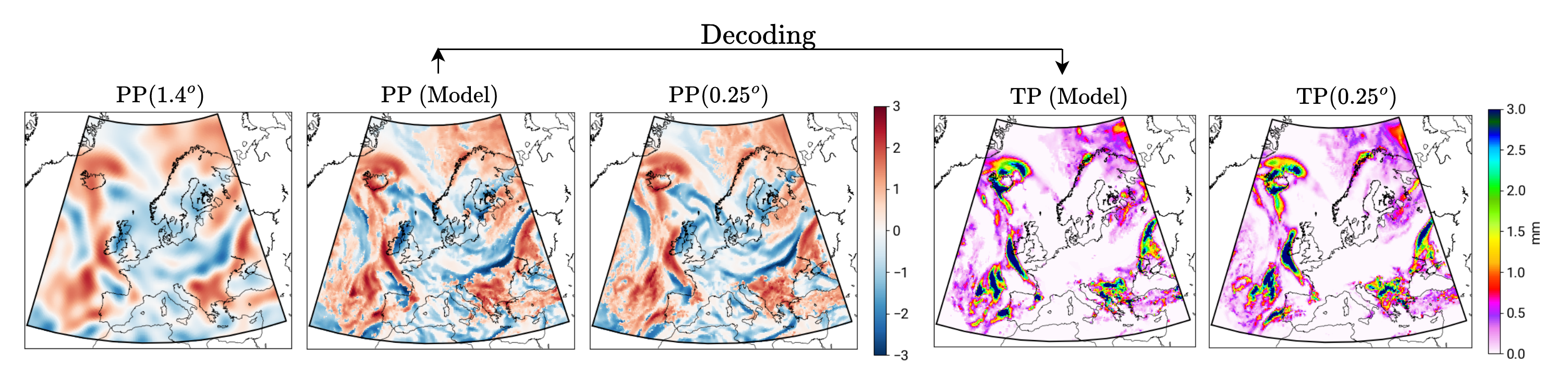}
  \caption{Visual assessment of downscaling for a snapshot. Downscaled PP is decoded to TP using the PP model decoder from figure \ref{fig:schematics}.}
  \label{fig:visual}
\end{figure}

\section{Results}
\label{sec:results}

\paragraph{Experimental protocol} 

The TP and VIMD variables in this study are accumulated over three-hour intervals. The downscaling model is trained on 30 years of ERA5 reanalysis data (1990-2020), with an additional 10 years used for testing and validation. The PP model is trained on one year of ERA5 data (2010). All experiments were conducted on T4 GPUs hosted on AWS.

\paragraph{Validation} 

\begin{figure}[h]
  \centering
  \includegraphics[width=1.01\textwidth]{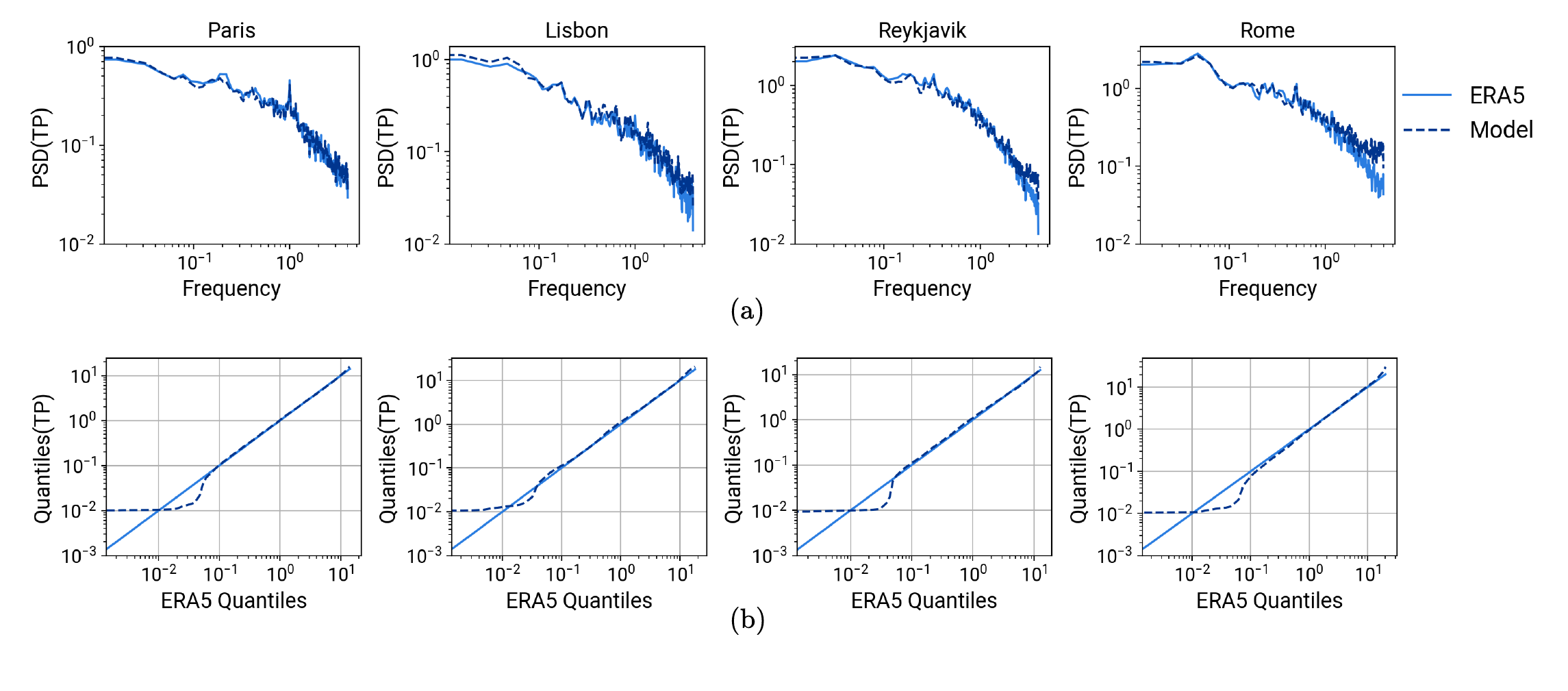}
  \caption{a) Power spectral density (PSD) and b) Q-Q plots.}
  \label{fig:psd}
\end{figure}

\begin{figure}[h]
  \centering
  \includegraphics[width=1.05\textwidth]{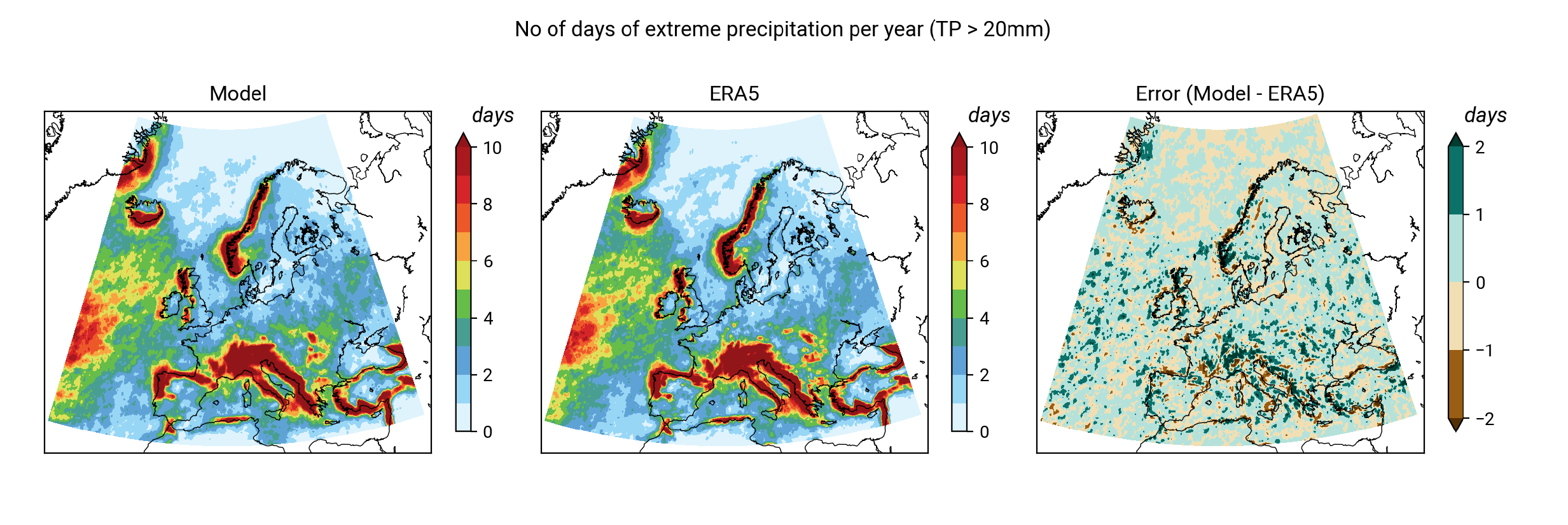}
  \caption{Number of days of extreme precipitation (TP > 20 mm).}
  \label{fig:bar}
\end{figure}

We evaluate the performance of our downscaling model over Europe by comparing the results in terms of visual agreement, power spectra at fixed locations, and local estimates of extreme precipitation. Figure \ref{fig:visual} provides a qualitative assessment of the predictions from our downscaling model, showing that the model successfully captures the fine-scale features (stochastic in nature), while preserving the large-scale structures. Figure \ref{fig:psd}a presents the temporal power spectrum and figure \ref{fig:psd}b displays quantile plots for major European cities. Additionally, figure \ref{fig:bar} illustrates the local estimates of the number of days of extreme precipitation (TP > 20 mm). Our results exhibit a strong agreement with the ERA5 dataset.

\section{Discussion}
\label{sec:discussion}
In this study we propose a machine learning-based approach for generating pseudo-precipitation which is a spatio-temporally smooth and continuous field derived from TP and VIMD. We demonstrate the advantages of using the pseudo-precipitation field as a robust alternative to precipitation, particularly in downscaling applications. The proposed methodology accurately estimates extreme precipitation and produces predictions that are consistent across the frequency spectrum when compared to ERA5. While this work primarily focuses on downscaling, the proposed pseudo-precipitation blending approach can also be applied to other statistical tasks, such as downscaling, debiasing and forecasting.

{
\small
\bibliography{refs}
}

\appendix

\section{Appendix}
\label{appendix}

\subsubsection{Training strategy}
\label{sec:training_method}

The pseudo-precipitation (PP) model uses an encoder-decoder architecture based on fully connected layers to process point-wise ERA5-resolution data, blending them into PP. A parametric study optimized the encoder and decoder depth and neurons per layer. To approximate a Gaussian distribution in PP, we apply a quantile loss that matches the quantiles of PP with those of a standard Gaussian. It is defined as:
\begin{equation}
    \textit{L}_{quant} = MSE(\textbf{q}(y_{pred}), \textbf{q}(y_{normal}))
    \label{eq:quantile}
\end{equation}

where \textbf{q} is a vector that contains the quantiles of the variable \textit{y}. Quantiles are computed using 4000 bins that span \(10^{-6}\) to \(1-10^{-6}\). A Mean Square Error (MSE) loss is also used to enforce the reconstruction of decoded TP from PP.: 
\begin{equation}
    \textit{L}_{total} = w_{quant}\textit{L}_{quant} + w_{rec}\textit{L}_{rec}
    \label{eq:total}
\end{equation}

where \(w_{quant}\) and \(w_{rec}\) are scalar weights for quantile and reconstruction loss terms, respectively. The combination of \(w_{quant} =20\) and \(w_{rec} =1\) achieved optimal PP blending and accurate TP reconstruction, producing a mean absolute error (MAE) of approximately \(10^{-6}\). 

\end{document}